\def\year{2020}\relax
\documentclass[letterpaper]{article} 
\usepackage{aaai20}  
\usepackage{times}  
\usepackage{helvet} 
\usepackage{courier}  
\usepackage[hyphens]{url}  
\usepackage{graphicx} 
\usepackage{multirow}
\usepackage{latexsym}
\usepackage{url}
\usepackage{subcaption}
\usepackage{array}
\usepackage{mathrsfs}
\usepackage{bm}
\usepackage{booktabs}
\usepackage{amsmath}
\usepackage{hhline}
\usepackage{framed}

\newcommand{\citet}[1]{\citeauthor{#1}\shortcite{#1}}
\newcommand{\citep}{\cite}

\title{Fact-aware Sentence Split and Rephrase with Permutation Invariant Training}
\author{Yinuo Guo\textsuperscript{\rm 1},  
Tao Ge\textsuperscript{\rm 2}, Furu Wei\textsuperscript{\rm 2} \\
\large 
\textsuperscript{\rm 1} Key Laboratory of Computational Linguistics\\ School of EECS, Peking University,\\
\textsuperscript{\rm 2} Microsoft Research Asia \\
gyn0806@pku.edu.cn\\
\{tage, fuwei\}@microsoft.com
}

\begin{document}
\maketitle
\begin{abstract}
Sentence Split and Rephrase aims to break down a complex sentence into several simple sentences with its meaning preserved. Previous studies tend to address the issue by seq2seq learning from parallel sentence pairs, which takes a complex sentence as input and sequentially generates a series of simple sentences. However, the conventional seq2seq learning has two limitations for this task: (1) it does not take into account the facts stated in the long sentence; As a result, the generated simple sentences may miss or inaccurately state the facts in the original sentence. (2) The order variance of the simple sentences to be generated may confuse the seq2seq model during training because the simple sentences derived from the long source sentence could be in any order.

To overcome the challenges, we first propose the Fact-aware Sentence Encoding, which enables the model to learn facts from the long sentence and thus improves the precision of sentence split; then we introduce Permutation Invariant Training to alleviate the effects of order variance in seq2seq learning for this task. Experiments on the WebSplit-v1.0 benchmark dataset show that our approaches can largely improve the performance over the previous seq2seq learning approaches. Moreover, an extrinsic evaluation on oie-benchmark verifies the effectiveness of our approaches by an observation that splitting long sentences with our state-of-the-art model as preprocessing is helpful for improving OpenIE performance. 
\end{abstract}

\begin{figure}[t]
\small
    \centering
    \includegraphics[width=.43\textwidth]{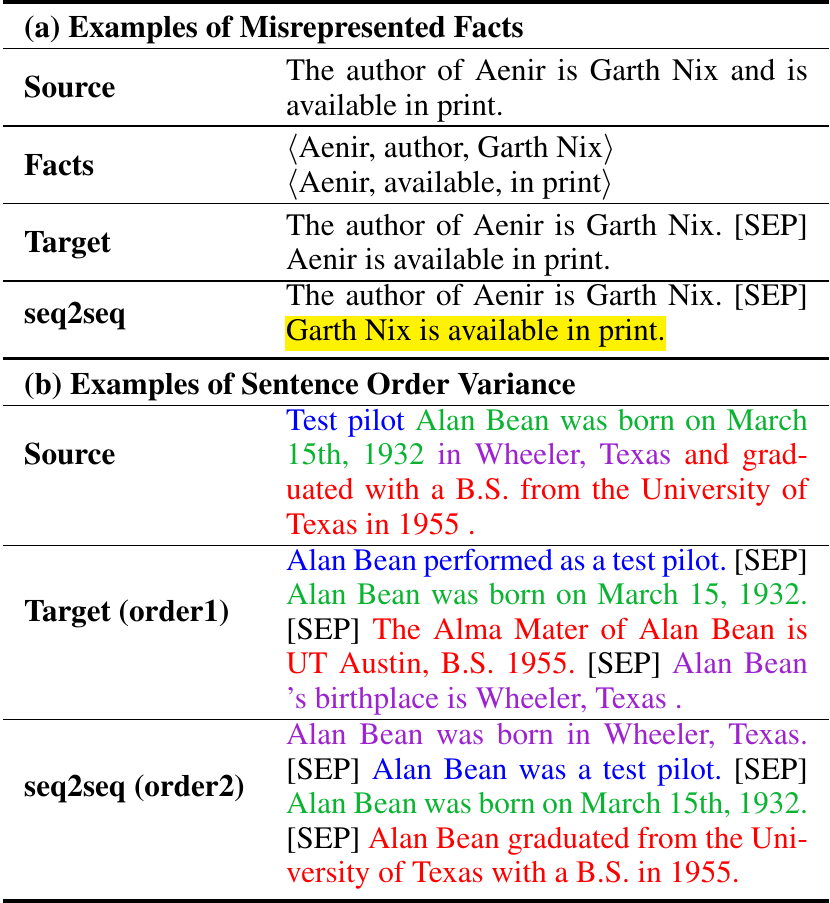}
    \caption{Examples of sentences generated by the seq2seq model. (a) illustrates misrepresented facts (yellow) in the generated results; (b) shows two split order of simple sentences derived from the source. The sentence order variance may confuse the model during training, tweaking the model against its previously learning patterns and affecting the stability of training. Best viewed in color.}
    \label{fig:examples}
\end{figure}

\begin{figure*}[!htbp]   
\small
 \center{\includegraphics[width=.92\textwidth]
 {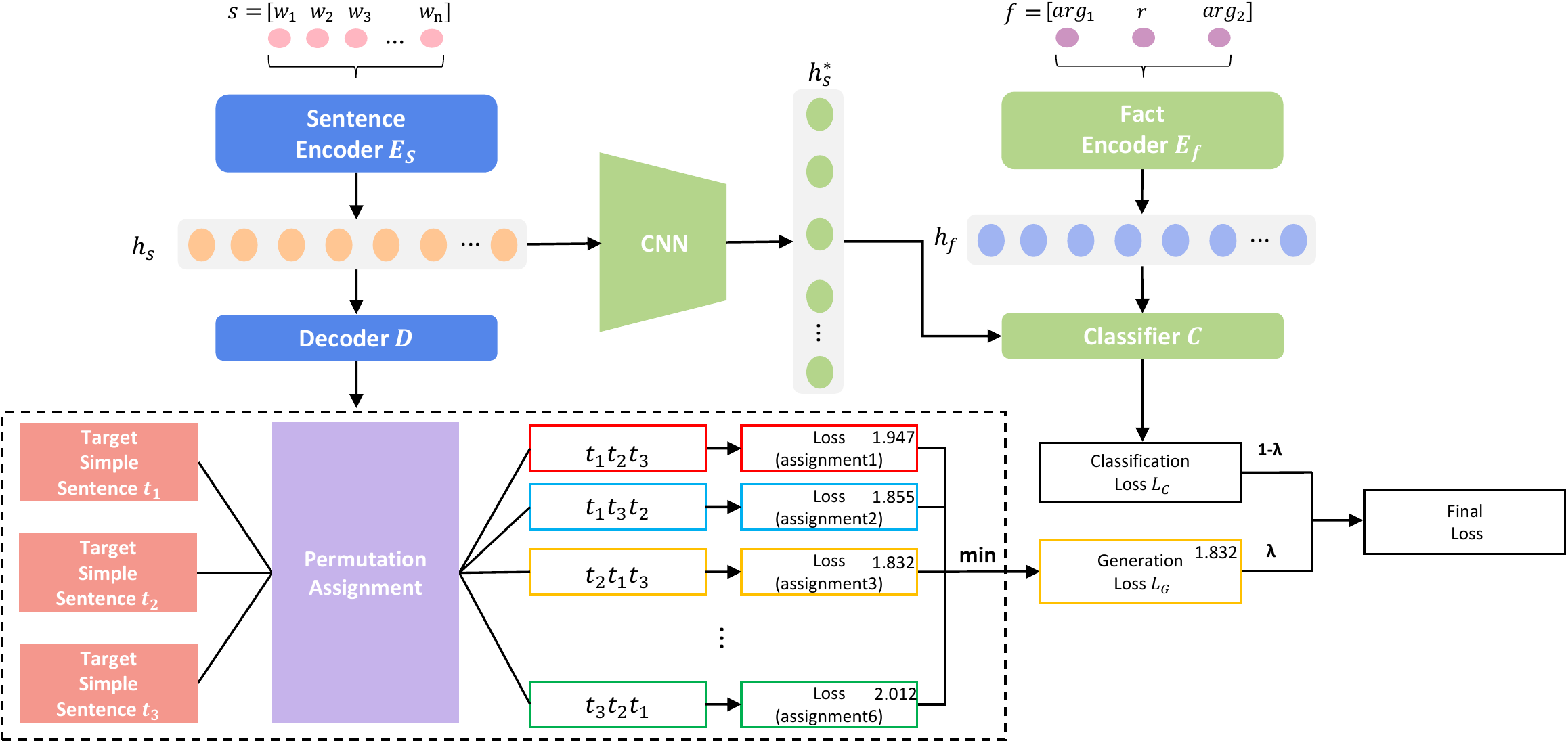}}        
\caption{The overview of our model architecture. The Sentence Encoder $E_s$ encodes the complex sentence ($w_1, w_2, w_3,\cdots,w_n$) into a representation vector $\boldsymbol{h_s}$, which is shared by two tasks: sentence split and rephrase, and fact classification. For sentence split and rephrase, Decoder $D$ takes $\boldsymbol{h_s}$ as input and generates a series of simple sentences. For fact classification, Fact Encoder $E_f$ encodes the fact $f=\langle arg_1, r, arg_2 \rangle$ into a representation vector $\boldsymbol{h_f}$, which will be fed into a classifier to determine whether $f$ is true or not based on a sentence $\boldsymbol{s}$. The model is trained in a multi-task learning fashion and the final loss is a weighted sum of the losses of the two tasks. Specifically, the loss of sentence split and rephrase task is obtained by the Permutation Invariant Training strategy which chooses the minimal loss among all the permutation assignments. Best viewed in color.}   
 \label{fig:overview}
 \end{figure*}
\section{Introduction}\label{sec:intro}

Long and complicated sentences prove to be a stumbling block for computers to process human languages. If complex text can be made shorter and simpler, sentences will become easier to understand, which will benefit a number of natural language processing (NLP) tasks such as Information Extraction.   

With this motivation, Sentence Split and Rephrase task was proposed~\cite{narayan2017split}, focusing on simplifying a complicated sentence by splitting and rephrasing it into simple sentences. Previous studies~\cite{narayan2017split,aharoni2018split,botha2018learning} tend to formulate this problem into a seq2seq learning framework, which takes the complex sentence as input and sequentially generates a series of simple sentences. Despite the certain effectiveness of these approaches, their limitations are obvious: (1) conventional seq2seq learning does not take into account the facts stated in the original sentence; As a result, the facts are likely to be misrepresented or missed in the generated simple sentences, as Figure~\ref{fig:examples}(a) illustrates. (2) The variance of the order of simple sentences to be generated may confuse the seq2seq model because the simple sentences derived from the long source sentence could be in any order. As Figure~\ref{fig:examples}(b) shows, if the sentence order in the target of a training instance is opposite to the way that the seq2seq model used to split, the model will be confused and tweaked against its previously learned patterns, affecting the stability of training.

To address these challenges, we propose a novel approach for this task with Fact-aware Sentence Encoding (FaSE) and Permutation Invariant Training (PIT). Fact-aware Sentence Encoding enhances the encoder of the seq2seq model by leveraging multi-task learning paradigm: it trains the sentence encoder not only for the Sentence Split and Rephrase task but also for judging whether a given fact is true or false based on the sentence, resulting in a fact-aware text encoder that is helpful for accurate sentence splitting; while Permutation Invariant Training, which has been widely used to solve the label permutation problem in the task of multi-talker speech separation, is introduced to find the best permutation of the simple sentences in the reference that yields the minimal loss for avoiding learning against the previously learned patterns.

Experiments on the WebSplit-v1.0 benchmark show that our proposed Fact-aware Sentence Encoding and Permutation Invariant Training can largely improve the performance over the conventional seq2seq models and achieve the state-of-the-art result. Moreover, an extrinsic evaluation carried out in oie-benchmark verifies that splitting long sentences with our approach as preprocessing is helpful for improving the performance of OpenIE.

Our contributions are summarized as follows:
\begin{quote}
\begin{itemize}
\item{We propose a novel Fact-aware Sentence Encoding approach that enables the model to learn facts from sentences for improving the accuracy of Sentence Split and Rephrase.}
\item{We are the first to introduce Permutation Invariant Training for Sentence Split and Rephrase task to alleviate the effects of sentence order variance in seq2seq learning.}
\item{We advance the state-of-the-art in Sentence Split and Rephrase task, and show that our Sentence Split and Rephrase model can effectively simplify long and complicated sentences to improve the performance of OpenIE.}
\end{itemize}
\end{quote}

\begin{table*}[!t]
\small
\centering
\begin{tabular}{l|p{9cm}}
\toprule[1pt]
\textbf{Source}
& A Wizard of Mars, by Diane Duane, has OCLC number 318875313 and ISBN number 978-0-15-204770-2.\\\hline
\specialrule{0em}{1pt}{1pt}
\multirow{2} *{\textbf{Positive}} & $\langle$A Wizard of Mars, author, Diane Duane$\rangle$\\
\specialrule{0em}{1pt}{1pt}
& $\langle$A Wizard of Mars, OCLC number, 318875313$\rangle$\\\hline
\specialrule{0em}{1pt}{1pt}
\textbf{Negative(Relation)}&{$\langle$A Wizard of Mars, has, Diane Duane$\rangle$}\\
\specialrule{0em}{1pt}{1pt}
\textbf{Negative(Arg)}
&$\langle$A Wizard of Mars, OCLC number variation, 978-0-15-204770-2$\rangle$ \\
\bottomrule[1pt]
\end{tabular}
    \caption{Examples of data instances for fact classification. Positive examples are obtained from WebNLG and RNNLG, while negative examples are derived by corrupting the facts in the positive instances in two ways: \textbf{Negative(Relation)} replaces the $relation$ in positive instances with a random relation in other facts; \textbf{Negative(Arg)} corrupts the positive instances by replacing one word in $arg1$ or $arg2$ with a random word in the source sentence. We construct negative examples with the two corruption strategies with the equal probability.}
\label{tab:fact_corupt}
\end{table*}

\section{Background}\label{sec:bg}
\subsection{Sentence Split and Rephrase}
As the examples shown in Figure~\ref{fig:examples}, Sentence Split and Rephrase task aims to split a complex sentence into several simple sentences with meaning preserved. Formally, given a complex source sentence $\boldsymbol{s}$, we expect to derive a list of simplified sentences $(\boldsymbol{t_1}, \boldsymbol{t_2}, \cdots, \boldsymbol{t_K})$, where $\boldsymbol{t_k}$ is one simplified sentence. The entries in the list $(\boldsymbol{t_1}, \boldsymbol{t_2}, \cdots, \boldsymbol{t_K})$ should rephrase different parts of $\boldsymbol{s}$, and should be faithful to the facts in $\boldsymbol{s}$. It is notable that the number of simplified sentences to be generated (i.e., $K$) is not given; in other words, a model has to determine $K$ by itself while splitting a complex sentence.

\subsection{Sequence-to-Sequence Learning}
Sequence-to-sequence (seq2seq) learning achieves tremendous success in many NLP tasks. A typical seq2seq model consists of an encoder and a decoder. The encoder first converts the input sentence $\boldsymbol{s}$ into dense vector representation $\boldsymbol{h}$, and then the decoder generates the target sentence $\boldsymbol{t}$ based on $\boldsymbol{h}$.

In practice, the encoder and the decoder are jointly trained by minimizing the negative log-likelihood of parallel source-target sentence pairs in the training set. At the inference time, an output sentence is generated in an auto-regressive manner by finding the most likely token sequence through beam search.

As previous work, when applied to Sentence Split and Rephrase task, a seq2seq model takes the complex sentence $\boldsymbol{s}$ as its input (i.e., source sentence). Its corresponding output $\boldsymbol{t}$ (i.e., target sentence) is its simplified sentences which are concatenated by a special token ``\textbf{[SEP]}'' indicating sentence boundaries, as the examples in Figure~\ref{fig:examples} shows.

\section{Methodology}
As introduced in Section \ref{sec:intro}, conventional seq2seq learning has obvious limitations for Sentence Split and Rephrase task: (1) it is not fact-aware and thus likely to misrepresent the facts in the simplified sentences; (2) it is sensitive to sentence order variance in the training data, resulting in an unstable training process. To overcome the limitations, we propose to improve seq2seq learning for this task by: (a) enhancing the text encoder by making it become fact-aware; (b) introducing Permutation Invariant Training to alleviate the issue of sentence order variance to stabilize the training process. The overview of our approach is illustrated in Figure \ref{fig:overview}. We will introduce the details of Fact-aware Sentence Encoding and Permutation Invariant Training in the remaining parts of this section.

\subsection{Fact-aware Sentence Encoding}\label{subsec:fae}
To make the model become fact-aware, we propose to enhance the encoder through a multi-task learning. As Figure \ref{fig:overview} shows, the sentence encoder $E_s$ is trained in a multi-task learning fashion: it is not only trained for the Sentence Split and Rephrase task, but also for judging (i.e., classifying) whether a given fact $f=\langle arg_1, r, arg_2 \rangle$ is true or false based on a sentence, where $r$ is the relation of the arguments $arg_1$ and $arg_2$. With the supervision of the fact classification objective during training, the sentence encoder $E_s$ will become more likely to accurately capture key facts from the sentence and yield fact-aware sentence encoding representation $\boldsymbol{h_s}$.

For training such a fact-aware seq2seq model for Sentence Split and Rephrase, we optimize for the following objective:

\begin{equation}\label{eq:multitask}
    L = \lambda L_G + (1 - \lambda)L_C 
\end{equation}
where $L_G$ is the loss for the task of Sentence Split and Rephrase, $L_C$ is the loss for the task of fact classification, and $\lambda$ is the hyper-parameter to control the weights for the two tasks.

As introduced in Section \ref{sec:bg}, the loss $L_G$ for Sentence Split and Rephrase can be optimized through maximum likelihood estimation (MLE) on parallel source (i.e., the complex sentence $\boldsymbol{s}$) - target (i.e., the concatenated simplified sentences $\boldsymbol{t_1}$[SEP]$\boldsymbol{t_2}\cdots$[SEP]$\boldsymbol{t_K}$) pairs\footnote{For simplicity, we call such pairs C-S pairs in the following parts of this paper.}.

For optimizing loss $L_C$ for fact classification, we first collect sentence-fact pairs $\langle \boldsymbol{s}, f \rangle$, where $\boldsymbol{s}$ is a sentence and $f=\langle arg_1, r, arg_2 \rangle$, from WebNLG~\cite{gardent-etal-2017-creating} and RNNLG~\cite{wen2016multi}. To construct negative examples for fact classification, we corrupt the fact $f$ by replacing one word in ${arg}_1$ or ${arg}_2$ with a random word in $\boldsymbol{s}$, or replacing $r$ with a random relation $r^*$ in other facts. Table~\ref{tab:fact_corupt} gives an example of the data instances for fact classification.

During training\footnote{In our experiments, the loss $L_G$ and $L_C$ in Eq (\ref{eq:multitask}) are the sum of the loss in the corresponding tasks in a batch.}, we use the equal number of training examples for the fact classification task and the Sentence Split and Rephrase task within a batch.

\begin{table*}
\small
\centering
\begin{tabular}{l|c|c|c}
\toprule[1pt]
  & BLEU& \#S/C & \#T/S  \\
  \midrule
 Reference& - & 2.5 & 10.9 \\
\midrule
\textsc{Source}  &57.2& 1.0 & 20.5  \\
\textsc{SplitHalf} & 55.7&  2.0 & 10.8 \\
\textsc{Lstm}(\textsc{AG18})~\cite{aharoni2018split} &25.5 & 2.3 & 11.8\\
\textsc{Lstm} (\textsc{Botha-WebSplit})~\cite{botha2018learning} & 30.5&  2.0 & 8.8  \\
\midrule
\multicolumn{4}{l}{\textsc{\textbf{Ours}}}\\
\midrule
\textsc{Lstm }(Baseline) &32.5 &  2.6 & 9.8 \\
\textsc{Lstm + Pit} & 34.7 & 2.5& 9.9 \\
\textsc{Lstm + FaSE} & 34.2 & 2.4 & 9.8\\
\textsc{Lstm + FaSE + Pit}& \textbf{37.3} &2.4 & 9.9 \\
\bottomrule[1pt]
\end{tabular}
\caption{Results on the WebSplit-v1.0. \textsc{Source} means directly taking the unmodified
source complex sentence as prediction. \textsc{SplitHalf} means deterministically
spliting a complex sentence into two equal-length
token sequences and appending a period to
the first one. \textsc{Lstm} (\textsc{AG18}) refers to the model in \citet{aharoni2018split}. \textsc{Lstm} (\textsc{Botha-WebSplit}) refers to the previous state-of-the-art proposed in~\cite{botha2018learning}. In our implementation, we observe that both \textsc{Pit} and \textsc{FaSE} can bring a profit to the \textsc{Lstm} baseline. And \textsc{Lstm + Pit + FaSE} which means the combination of \textsc{Pit} and \textsc{FaSE} achieves a new state-of-the-art.}\label{tab:lstm_res}
\end{table*}

\subsection{Permutation Invariant Training}\label{subsec:pit}
As introduced in Section \ref{sec:intro}, the issue of sentence order variance affects seq2seq learning for Sentence Split and Rephrase and consequently makes the model training very unstable. Inspired by the successful practice of Permutation Invariant Training to solve the label permutation issue in the task of multi-talker speech separation, we propose to introduce Permutation Invariant Training in Sentence Split and Rephrase task.

The key idea of Permutation Invariant Training is to find the best permutation of the simplified sentences in the reference that yields the minimal loss to update the model. To make it easy to understand, we use the illustration inside the dashed rectangular in Figure \ref{fig:overview} to visualize this process. In Figure \ref{fig:overview}, the complex sentence $\boldsymbol{s}$ corresponds to three simple sentences $\boldsymbol{t_1}$, $\boldsymbol{t_2}$ and $\boldsymbol{t_3}$. We first enumerate all permutation assignments (total number of $3!=6$) from the three simple sentences, and then compute the loss for each assignment. Finally, we choose the permutation with the minimal loss (i.e., assignment 3 in Figure \ref{fig:overview}) to update the model. In this way, the issue of sentence order variance can be alleviated, helping the model avoid learning against its previously learned patterns and stabilize the training process.

\subsection{Inference}
At inference time, our model behaves totally the same as conventional seq2seq models, which takes the complex sentence as the input, and generates concatenation of simple sentences in an auto-regressive manner by finding the most likely token sequence through beam search.

\section{Experiments}
\subsection{Data}
We conduct experiments on the WebSplit-v1.0 corpus, which is a benchmark dataset to compare the performance of models for Sentence Split and Rephrase task. Its training set contains approximately 1.3 million C-S pairs; while its validation and test sets contain about 4,000 complex sentences, each of which is equipped with multiple references.

For Fact-aware Sentence Encoding with multi-task learning, we construct a fact classification dataset based on the method mentioned in Section \ref{subsec:fae}. The resulting dataset contains 94,930 training samples where half are positive and the other are negative samples constructed by fact corruption. 

\subsection{Evaluation Setting}\label{subsec:setting}
\subsubsection{Model Configuration}
As previous work \cite{aharoni2018split,botha2018learning}, we use 1-layer LSTM (512 hidden units) encoder-decoder model with attention and copy mechanism~\cite{gu2016incorporating} as our seq2seq model. For fact classification, as illustrated in Figure \ref{fig:overview}, we first use a CNN layer with Relu activation and  max-pooling to generate a fixed length vector $\boldsymbol{h_s^*}$ from $\boldsymbol{h_s}$, then concatenate $\boldsymbol{h_s^*}$ with the fact vector $\boldsymbol{h_f}$ encoded by a fact encoder $E_f$ which is also a CNN layer with the same configuration: filter size $n={3, 4, 5}$ and filter number 24 for each size. To train the models, we use all the words (7k) in the training data as the vocabulary for both source and target. We use the Adam optimizer with a learning rate of 0.0005 with 8000 warmup steps. The training process lasts 30 epochs with the batch size of 64. During inference, the beam size is set to 12.

\subsubsection{Evaluation Metrics}
Following the prior research in the Websplit-v1.0 benchmark~\cite{botha2018learning,aharoni2018split}, we report the sentence-level \textsc{BLEU}\footnote{The sentence-level BLEU score calculated by the evaluation script in https://github.com/roeeaharoni/sprp-acl2018} and length based statistics to quantify splitting.
\begin{table*}[!htbp]
\small
\centering
\begin{tabular}{l|c|c|c}
\toprule[1pt]
  & BLEU & \#S/C & \#T/S  \\
 \midrule
 Reference& - & 2.5 & 10.9 \\
 \midrule
\textsc{Source}  &57.2&   1.0 & 20.5  \\
\textsc{SplitHalf}& 55.7&   2.0 & 10.8 \\
\textsc{DisSim}~\cite{niklaus2019transforming} &59.7 & 2.7& 9.2\\
\textsc{Botha-both}~\cite{botha2018learning} & 60.1 &  2.0 & 11  \\
\midrule
\multicolumn{4}{l}{\textsc{\textbf{Ours}}}\\
\midrule
\textsc{Transformer} &69.7& 2.3& 10.1 \\
\textsc{Transformer+ Pit}  &70.5&2.5& 9.9 \\
\textsc{Transformer + FaSE} & 70.7  &2.3 & 9.8 \\
\textsc{Transformer+ FaSE + Pit}& \textbf{71.0}  & 2.4 & 9.8 \\
\bottomrule[1pt]
\end{tabular}
\caption{Results of pre-training experiments on the WikiSplit.
\textsc{Transformer} denotes basic pre-trained Transformer model and \textsc{Transformer+Pit} additionally incorporates the PIT strategy into training. \textsc{FaSE} denotes the single Fact-aware Sentence Encoding proposed in~\ref{subsec:fae}
, and \textsc{FaSE+Pit} represents the combination of FaSE and PIT. Note that the FaSE and the PIT are applied at the fine-tune stage, we just pre-train the seq2seq (Transformer) model on the WikiSplit.}\label{tab:boosting_res}
\end{table*}
\subsection{Results}
For experiment results presented in Table~\ref{tab:lstm_res}, we see that our proposed methods largely improve the performance compared to the conventional seq2seq learning approaches. Surprisingly, the combination of PIT and FaSE (\textsc{Pit + Fase}) achieves 37.3 BLEU score, outperforming the best previously reported model trained on the WebSplit-v1.0 (\textsc{Botha-WebSplit}) by more than 6.8 BLEU, and gains 4.8 BLEU over our stronger baseline. As an ablation study, we observe that the single FaSE brings 1.7 BLEU improvement compared to the baseline model training without facts. This demonstrates the effectiveness of the FaSE which helps the text encoder explicitly capture more fact information from the source sentence, leading to a higher accuracy for the sentence splitting. On the other hand, the PIT strategy contributes 2.2 BLEU improvement to LSTM (Baseline). The significant improvement proves that PIT can help the model better learn to split and rephrase without being distracted by the trivial sentence order issue.

 \begin{table}[!tbp]
 \small
\centering
\begin{tabular}{l|c|c|c}
\toprule[1pt]
  & BLEU & \#S/C & \#T/S  \\\hline
\textsc{Lstm-Max} &  $23.9$ &$2.7$ & $9.6$ \\
\textsc{Lstm-Random} &$25.9\pm1.2$ &$2.6\pm0.3$& $9.5\pm0.1$ \\
\textsc{Lstm-Min (PIT)}& \textbf{34.7} & $2.5$  & $9.7$ \\
\bottomrule[1pt]

\end{tabular}
\caption{Results of experiments on sentence order variance problem. \textsc{Max}, \textsc{Min} denotes choosing the permutation with the max loss, min loss separately. And \textsc{Random} represents randomly selecting one permutation. Note that \textsc{Min} here is exactly the PIT strategy. For the ``Random" setting, we launch the training process with different seeds five times and then calculate their mean and standard deviation. }\label{tab:pit}
\end{table}
 \subsubsection{Order Variance Exploration} Additionally, we conduct experiments to investigate to what extent sentence order variance influences the performance of the Sentence Split and Rephrase task. Specifically, the experiments are conducted in three settings: \textsc{Max}, \textsc{Random} and \textsc{Min}. \textsc{Max} and \textsc{Min} denote choosing the permutation with the max loss and min loss, respectively. And \textsc{Random} represents randomly selecting one permutation. Note that \textsc{Min} here is exactly the PIT strategy introduced in~\ref{subsec:pit}.  
 
 As shown in Table~\ref{tab:pit}, it is obvious that \textsc{Min} (PIT) outperforms the others by a substantial margin and \textsc{Max} performs the worst as expected. Furthermore, it is interesting to observe the instability of the \textsc{Random} according to the high standard deviation of 1.2. Apparently, the conventional seq2seq models are sensitive to the uncertainties brought by the sentence order variance, because they will be confused and tweaked against its previously learned patterns.
 
\subsubsection{{Large-scale Data Boosting}} 
 To further verify the performance of our methods on large-scale datasets, we incorporate the WikiSplit~\cite{botha2018learning} data for pre-training. The WikiSplit contains approximately 1.0 million C-S pairs on a rich and varied vocabulary mined from the Wikipedia edit histories. When using the WikiSplit for pre-training, all texts are encoded with byte-pair-encoding~\cite{sennrich2015neural} through sub-word units, which has a shared source-target vocabulary of about 33,000 words. We choose the base Transformer~\cite{vaswani2017attention} architecture equipped with the copy mechanism~\cite{gu2016incorporating}. The pre-training process on the \textsc{WikiSplit} lasts 10 epochs with the max tokens 2,048 for each batch. And the fine-tuning process on the \textsc{WebSplit} lasts 30 epochs with the batch size of 32 sentences. We use Adam to optimize the model with a learning rate 5e-4 at the pre-training stage and 5e-5 at the fine-tuning stage. Other parameters are consistent with the settings described in Section~\ref{subsec:setting}.

As shown in Table~\ref{tab:boosting_res}, due to its large amount and great diversity, the \textsc{WikiSplit} can be used to bootstrap the full model. And after applying our approaches to the stronger baseline, a new state-of-the-art performance of 71.0 BLEU score is obtained, which outperforms the baseline by 1.3 BLEU score. Furthermore, this state-of-the-art model is adopted to investigate whether our model can benefit downstream task OpenIE in Section~\ref{sec:openie}.

\begin{table}[t]
\small
\centering

\begin{tabular}{l|c|c|c|c}
\toprule[1pt]
  & G & M & S & avg.  \\\hline
\textsc{Simple Refernce} & $4.92$ &$4.64$ & $1.67$ & $3.74$ \\
\midrule
\textsc{Botha}~\cite{botha2018learning}& $3.95$& $3.77$  & $0.76$ & 2.83\\
\textsc{DisSim}~\cite{niklaus2019transforming} &$4.27$ &$3.65$& $1.1$ & 3.01\\
\textsc{Fase+Pit}& \textbf{4.78} & \textbf{4.4}  & \textbf{1.69} & \textbf{3.62} \\
\bottomrule[1pt]
\end{tabular}
\caption{Human evaluation results on a random sample of 50 sentences on the WebSplit-v1.0.}\label{tab:human}
\end{table}
 
\begin{figure*}[t]
\centering	
    
    \begin{subfigure}{.45\textwidth}
    	\centering
    	\includegraphics[width=\textwidth]{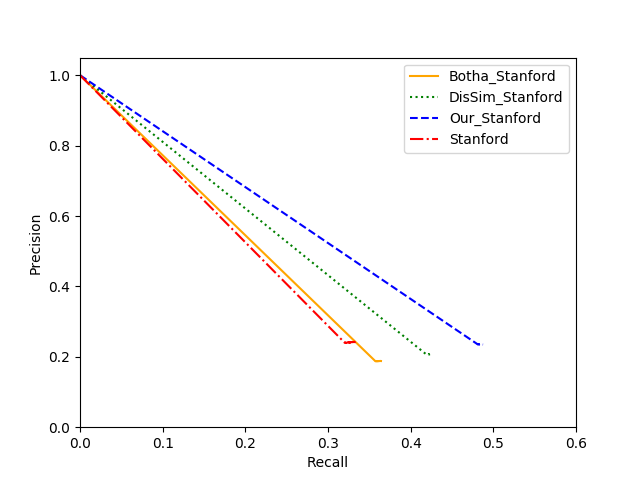}
    	\caption{indomain}
    \end{subfigure}
    \begin{subfigure}{.45\textwidth}
    	\centering
    	\includegraphics[width=\textwidth]{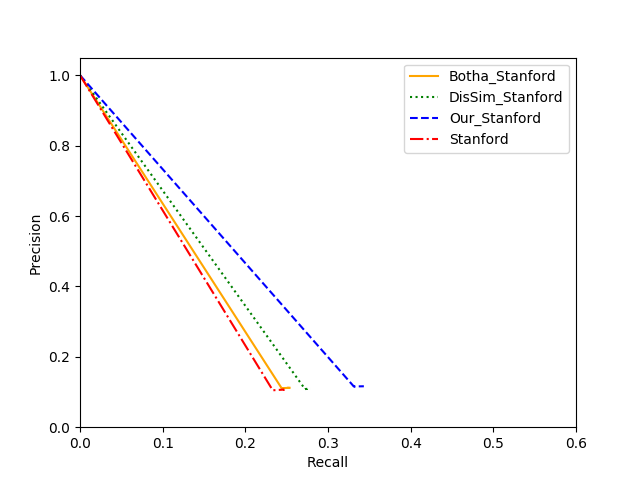}
    	\caption{oie-benchmark}
    \end{subfigure}

\caption{Performance of the Stanford \textsc{OpenIE} system with or without ``Split and Rephrase" model as a preprocessing step. ``DisSim\_Stanford" and ``Botha\_Stanford" denote the \citet{niklaus2019transforming} and \citet{botha2018learning} separately in the preprocessing step. ``Our\_Stanford" denotes the model in this paper.}
\label{fig:stanford_res}
\end{figure*}

\subsubsection{Human Evaluation}
Human evaluation is conducted on a subset of 50 randomly sampled sentences by two non-native, but fluent-English speakers who rated each C-S pair from 3 aspects: grammaticality~(G), meaning preservation~(M) and structural simplicity~(S). Annotation guidelines are described in the Appendix A  in~\citet{niklaus2019transforming}. The inter-annotator agreement was calculated using the Pearson correlation, resulting in rates of 0.64~(G), 0.66~(M) and 0.70~(S). As shown in Table~\ref{tab:human}, our model is rated higher than other baselines in all dimensions and performs close to the golden simple references. The results suggest that our model can transform the complex sentences into several simple sentences that achieve a high level of grammaticality and preserve the meaning of the input.

\subsection{Extrinsic Evaluation On OpenIE} \label{sec:openie}
To further validate the effectiveness of our model on downstream tasks, we use the state-of-the-art model as the preprocessing tool for the Stanford \textsc{OpenIE} \cite{angeli2015leveraging} system. More specifically, we firstly use our model to transform the input complex sentence into several simple sentences, and then send them to the Stanford \textsc{OpenIE} system for information extraction. 

We conduct experiments on an in-domain evaluation set extracted from the WebSplit-v1.0~\cite{narayan2017split} and the oie-benchmark~\cite{stanovsky2016creating} separately. Table~\ref{tab:ie_statistic} gives statistics on the in-domain evaluation set and the oie-benchmark. We compare the performance w/ and w/o our model applied as a preprocessing step. The evaluation metrics are precision and recall.
\begin{table}[!htbp]
\small
    \centering
    \begin{tabular}{|l|c|c|c|}
    \hline
     &sentences &relations & avg (\#R/S) \\\hline
      in-domain&930& 2,585 & 2.8\\\hline
      oie-benchmark&3,200& 10,359& 3.2\\\hline
    \end{tabular}
    \caption{Corpus statistics in terms of sentences, relations and average number of relations per sentence (avg) on the in-domain evaluation set and the oie-benchmark.}
    \label{tab:ie_statistic}
\end{table}

\begin{table*}[t]
\small
\centering
\begin{tabular}{|m{6.5cm}|m{7cm}|}
\hline
\textbf{Input (Complex Sentence)} & \textbf {Output (Simple Sentences)}\\\hline
A different judge then ordered the case reviewed by a higher court.&  A different judge then ordered the case. \textbf{[SEP]} The case was reviewed by a higher court.\\\hline
From 1909 to 1912, the Miami Canal was dug, bypassing the rapids at the head of the North Fork. & Miami Canal was dug from 1909 to 1912. \textbf{[SEP]} The Miami Canal bypassed the rapids at the head of the North Fork.\\\hline
Green is also a significant color, symbolizing the union of the three colors. & Green is a significant color. \textbf{[SEP]} Green symbolizes the union of the three colors. \\\hline
Wynne married Mary Ellen McCabe, daughter of a wealthy construction contractor. & Wynne married Mary Ellen McCabe. \textbf{[SEP]} Mary Ellen McCabe is the daughter of a wealthy construction contractor. \\\hline
As a successful young businessman in Detroit, Chandler supported the Underground Railroad. & Chandler was a successful young businessman in Detroit. \textbf{[SEP]} Chandler supported the Underground Railroad .\\\hline
He was a member of the European Convention, which drafted the text of the European Constitution that never entered into force. & He was a member of the European Convention. \textbf{[SEP]} European Convention drafted the text of the European Constitution. \textbf{[SEP]} The European Constitution never entered into force.\\\hline
\end{tabular}

\caption{Examples of sentences in the oie-benchmark preprocessed by our model.}\label{tab:oie-ex}
\end{table*}

\subsubsection{In-domain Evaluation}
Here we describe the construction of the in-domain evaluation set. 
The WebSplit-v1.0~\cite{narayan2017split} maps one complex sentence to a series of simple sentences which convey the same meaning, together with RDF triples that describe their semantics. Taking advantage of the provided RDF triples, we produce an in-domain evaluation set including all complex sentences in the WebSplit-v1.0 test data. For each complex sentence, not only its own RDF triples are included, but also the RDF triples that are related to its simple sentences. The WebSplit-v1.0 test data consists of 930 unique complex sentences mapping to 43,968 simple sentences. In total, we extract 2,585 relation triples as the in-domain evaluation set.

 \subsubsection{Oie-benchmark Evaluation}
 The experiments above mainly focus on the in-domain setting. To evaluate the generalization of our model, we further conduct experiments on the out-domain setting. We choose the popular large benchmark oie-benchmark~\cite{stanovsky2016creating} which contains 3,200 sentences with 10,359 extractions for evaluation.

\subsubsection{Analysis} We use the script in oie-benchmark\footnote{https://github.com/gabrielStanovsky/oie-benchmark} to evaluate the precision and recall of different models. Figure~\ref{fig:stanford_res} shows the precision-recall curves of the in-domain and oie-benchmark experiments. We also make a comparison among using our model and the other two models~\cite{botha2018learning,niklaus2019transforming} in the preprocessing step.  It is observed that our model can provide a strong boosting for the Stanford OpenIE system. We think the key reason is that our model turns the complex sentence into several faithful simple sentences and thus solves the OpenIE task in a divide-and-conquer manner, which makes the relation extraction process easier. We also give some examples in Table~\ref{tab:oie-ex} which shows that sentences in a new domain are also well processed thus proving the good generalization of our model. Clearly, the results taken together demonstrate the effectiveness of our approaches.

\section{Related Work}
\subsubsection{Split and Rephrase}

More recently, \citet{narayan2017split} proposes a new sentence simplification task dub ``Split and Rephrase'' and meanwhile introduces the WebSplit corpus, a dataset of over one million tuples which maps a single complex sentence to a sequence of simple sentences. They also release a series of seq2seq models trained on the WebSplit. Since then, this newly raised task has received a lot attention and various efforts have been made on data-driven approaches. \citet{aharoni2018split} then improve the WebSplit by reducing overlap in the data splits and presenting the neural model equipped with the copy mechanism. Even so, the encoder-decoder based models still perform poorly and \citet{botha2018learning} discovers that the sentences in the WebSplit corpus contain fairly unnatural linguistic expressions on a small vocabulary. Then they introduce a new training data mined from the Wikipedia edit histories, which includes a rich and varied vocabulary naturally expressed sentences and their extracted splits. However, the WikiSplit is limited to provide strong supervision because targets in it are not always the golden split of the source.
Different from the former data-driven approaches, \citet{niklaus2019transforming} proposes a recursive method using a set of hand-crafted transformation rules to split and rephrase complex English sentences into a novel semantic hierarchy of simplified sentences, whereas, rule-based approaches tend to be extremely labor intensive to create and have very poor generalization.

 \subsubsection{Permutation Invariant Training (PIT)}
Permutation Invariant Training \cite{yu2017permutation,chen2017deep,kolbaek2017multitalker} is a training strategy which effectively solves the long-lasting label permutation problem for deep learning based speaker independent multi-talker speech separation. The label permutation problem, also known as the label ambiguity problem, comes from the assignment between the separated frames and the golden frames. During training, the error between the clean magnitude spectra $a_{1,i}$ and $a_{2,i}$ and their estimated counterparts $\hat{a}_{1,i}$ and $\hat{a}_{2,i}$ needs to be computed. However, the one-to-one correspondence between the separated frames and the golden frames is unknown. To solve this issue, the PIT first determines the best output-target assignment and then minimizes the error given the assignment. This strategy, which is directly implemented inside the network structure, elegantly solves the problem which has prevented progress on deep learning based techniques for speech separation.

\section{Conclusion}
In this paper, we present a novel framework for the Sentence Split and Rephrase task consisting of Fact-aware Sentence Encoding (FaSE) and Permutation Invariant Training (PIT). To address the limitations of the conventional seq2seq methods for this task, the FaSE leverages the multi-task learning paradigm to make the text encoder more fact-aware and thus generates faithful simple sentences, and the PIT strategy alleviates the issue brought by sentence order variance to stabilize the training process. Extensive experiments demonstrate that both FaSE and PIT can bring a profit to the baseline model, and that their combination further improves the result and achieves the state-of-the-art performance on the WebSplit-v1.0 benchmark. Also, the profits on the OpenIE task verify that using our model as a preprocessing tool can facilitate this task and improve the performance, demonstrating the potential of our work for benefiting downstream NLP tasks.

\section*{Acknowledgments}
We thank the anonymous reviewers for their valuable comments. We want to thank Ke Wang in Microsoft STCA for the discussion and many constructive suggestions about PIT for this work. Specially, we thank Prof. Junfeng Hu for his guidance and support during the first author's academic career.

\bibliography{AAAI-GuoY.3736}
\bibliographystyle{aaai}
\end{document}